\documentclass[conference]{IEEEtran}

\usepackage[utf8]{inputenc}
\usepackage[pdftex]{graphicx}
\usepackage{subfig}
\usepackage{enumitem,kantlipsum}
\usepackage{svg}

\usepackage[a4paper, total={6in, 8in}]{geometry}

\graphicspath{{./figures/}}

\usepackage{pgffor}

\title{XAI for Self-supervised Clustering of Wireless Spectrum Activity}

\author{\IEEEauthorblockN{Ljupcho Milosheski, Gregor Cerar, Bla\v{z} Bertalani\v{c}, Carolina Fortuna and Mihael Mohor\v{c}i\v{c}}\\
\textit{Jozef Stefan Institute, Ljubljana, Slovenia}\\
\{ljupcho.milosheski, gregor.cerar, blaz.bertalanic, carolina.fortuna, miha.mohorcic\}@ijs.si}

\begin{document}

\maketitle

\begin{abstract}
The so-called black-box deep learning (DL) models are increasingly used in classification tasks across many scientific disciplines, including wireless communications domain. In this trend, supervised DL models appear as most commonly proposed solutions to domain-related classification problems. 
Although they are proven to have unmatched performance, the necessity for large labeled training data and their intractable reasoning, as two major drawbacks, are constraining their usage. The self-supervised architectures emerged as a promising solution that reduces the size of the needed labeled data, but the explainability problem remains.
In this paper, we propose a methodology for explaining deep clustering, self-supervised learning architectures comprised of a representation learning part based on a Convolutional Neural Network (CNN) and a clustering part. For the state of the art representation learning part, our methodology employs Guided Backpropagation to interpret the regions of interest of the input data. For the clustering part, the methodology relies on Shallow Trees to explain the clustering result using optimized depth decision tree. Finally, a data-specific visualizations part enables connection for each of the clusters to the input data trough the relevant features. We explain on a use case of wireless spectrum activity clustering how the CNN-based, deep clustering architecture reasons.

\end{abstract}

\begin{IEEEkeywords}
XAI, explainable AI, self-supervised learning, spectrum sensing, transmission classification
\end{IEEEkeywords}

\section{Introduction}

In the last ten years, countless research works proposed employing deep learning models for a wide variety of classification and regression tasks in various fields that depend on signal processing. These, so-called black-box models, are shown to have superior performance both on publicly available benchmarks as well as on application-specific datasets. The field of wireless communication technologies is no exception to this trend. There are various field-specific implementations dedicated to the processing of spectral data of wireless transmissions for classification purposes. However, training a classifier based on a deep learning (DL) architecture requires large labeled data. Unlike for other fields such as image processing, big labeled datasets are not available and the labeling of spectral data is proved to be inaccurate and expensive to obtain \cite{gale2020automatic}. For such applications, self-supervised deep learning (SSDL) appears a promising approach \cite{milosheski2022self} with the capability of automatic processing of large amounts of unlabeled data for representation learning, and using the representations for downstream tasks.

Deployment of any technology for classification tasks in critical sectors, like communications or power system, implies the necessity of reasoning behind the decisions provided by that technology. And this is one of the main disadvantages of the black-box models. The number of parameters of a typical DL classification architecture can vary from few to several hundreds of millions, which makes the models intractable in their reasoning. As a consequence, techniques known under the phrase eXplainable AI (XAI) are gaining momentum \cite{arrieta2020explainable} in recent years in parallel to the black-box models. Their main purpose is to provide explainability and interpretability of the AI algorithms' decisions. The development of XAI models is envisioned as crucial and to have key impact on the next generation wireless networks \cite{guo2020explainable}, \cite{wang2021explainable}.

In this paper we demonstrate our work towards explaining self-supervised architectures for wireless spectrum data processing applications, based on Convolutional Neural Network (CNN) for representation learning and a clustering algorithm for supervision. Although there are some recent existing works towards explaining the self-supervised models, they are mostly focused on the CNN part \cite{gur2021visualization}, \cite{basaj2021explaining}. However, clustering is inseparable part of this type of methods in the training phase and should also be considered in the explanation of the reasoning. Considering this, we developed a new technique that relies on the Guided Backpropagation for CNN and Shallow Trees for clustering explainability, and complemented it with spectrum-data-specific visualizations. Thus, we provide an end-to-end connection of the input data content, through the CNN activation to the relevant representation features and the clustering output. To the best of our knowledge, there is no existing approach for comprehensive explainability of the SSDL architectures as a single entity (i.e., CNN + clustering), for the applications of spectrum data processing in the wireless communications domain.

The main contribution of this work is a new technique for explaining SSDL architectures comprised of CNN-based representation learner and clustering-based supervision applied to clustering of wireless transmissions using spectrograms. The approach is model-dependent (constrained to specific type of deep clustering architectures), but also general in the sense that it can be applied regardless of the type of CNN and clustering algorithm used.

The rest of this paper is organized as follows. Section~\ref{Sec:relatedwork} provides related work. Section~\ref{Sec:architecture} introduces the SSDL architecture and the dataset used in the study. Section~\ref{sec:methodology} proposes the methodology for explaining the architecture used for clustering of wireless transmissions while Section~\ref{Sec:results} provides results and discussion. Finally, Section~\ref{Sec:conclusions} concludes the paper.

\begin{figure}[!t]
    \centering
    \includegraphics[width=\columnwidth]{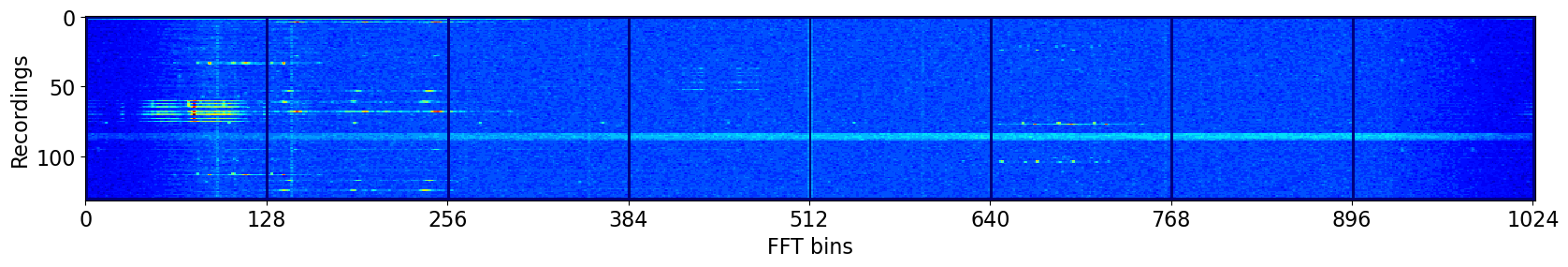}
    \caption[width=\textwidth]{Sample of 8 spectrogram segments from the data.}
    \label{fig:SampleData}
    \vspace{-5mm}
\end{figure}

\vspace{-2mm}
\section{Related work}
\label{Sec:relatedwork}

In the following subsections we provide a brief overview of the most relevant related work separately for CNN and clustering explainability.

\subsection{CNN explainability}

In \cite{Zhang2021ExplainabilityOS}, a SSDL architecture is used for representation learning for medical ultrasound video, where authors estimate the quality of the learned representation by the quality of the clustering. They use standard clustering metrics, i.e., Silhouette, Compactness and Uniqueness. The approach is specific to the application and difficult to extend to other domains because it is based on the assumption that the data contains meaningful captures made purposely by the domain expert (a physician). This is not the case in most problems for which data acquisition is performed without any human supervision or control. 

In \cite{gur2021visualization} authors' focus is on visualization of the network activation for a given class is presented . They achieve significant improvement on the visualizations compared to the existing solutions, but again there is no explainability of the pretext training process.

An innovative approach of applying a natural language processing probing task to image processing is proposed in \cite{basaj2021explaining}. Probing requires pretrained CNN that could provide visual words (superpixels) representing the analogue of words in a sentence. However, the existing models for processing spectrum data are case specific, utilizing data acquired by different types of devices, far less universal than RGB images and hence difficult to generalize. Another issue is the poor amount of content in the spectrogram images compared to the RGB images, which could result in a very small set of visual words.

Although the mentioned works are all worthy approaches towards explaining of the CNN-extracted features, they lack the interpretation of the pretext (clustering) task of the SSDL.

\subsection{Clustering explainability}

Building Predictive Clustering Trees (PCT) \cite{blockeel2000top} has been a well known approach for many years. It provides explainable models that allow finding explicit relations from the formed clusters to the feature space. This is especially important when the feature space is multidimensional and there are many existing clusters. In such cases, 2D and 3D projection visualizations of, for e.g., 20-dimensional data, do not provide enough insight for clarification of the groups of samples that are part of particular clusters obtained by a clustering algorithm (e.g. K-means). Recent publication with a comprehensive theoretical analysis of the usage of trees for explaining clustering algorithms is \cite{dasgupta2020explainable}.

Building a clustering tree while optimizing the inter-cluster distances as a splitting metric and the variance for stopping criteria can result in a very deep tree structure, thus leading to a model with low explainability. Recently, a different approach for building PCTs was proposed in \cite{laber2022shallow} where the depth of the tree is also optimized (minimized) in the tree-building algorithm. We take this algorithm as a baseline and complement it with spectrum-data-specific visualizations for the purpose of providing explainability of the clustering part of the SSDL architecture. 
\section{SSDL architecture and dataset}
\label{Sec:architecture}

For the demonstration of the proposed technique we used real-world data and the DeepCluster architecture \cite{caron2018deep} adapted for the wireless communications domain according to \cite{milosheski2022self}.

\subsection{Data} \label{sec:Data}
The dataset used for the analysis consists of fifteen days of radio spectrum measurements acquired at a sampling rate of 5 power spectral density (PSD) measurements per second using 1024 FFT bins in the 868\,MHz license-free (shared spectrum) frequency band with a 192\,kHz bandwidth. The data was acquired in the LOG-a-TEC testbed. Details of the acquisition process and a subset of the data can be found in \cite{vsolc2015low}. The acquired data has a matrix form of $1024 \times N$, where $N$ is the number of measurements over time. 
The segmentation of the complete data-matrix into non-overlapping square image-like spectrograms along time and frequency (FFT bins) is realized for a window size $W=128$. An example of such segmentation  containing 8 square images is shown in Figure~\ref{fig:SampleData}, corresponding to the image resolution of 25.6 seconds (128 measurements taken at 5 measurements per second) by 24 kHz. The window size is large enough to contain any single type of activity and small enough to avoid having too many activities in a single image, having in mind also computational cost. Dividing the entire dataset of 15 days using $W=128$ and zero overlapping, produces $423,904$ images of $128 \times 128$ pixels whose values are scaled to the range [0,1].

\subsection{Architecture} \label{sec:Architecture}

\begin{figure}[!t]
    \noindent\makebox[\columnwidth]{\includegraphics[width=\columnwidth]{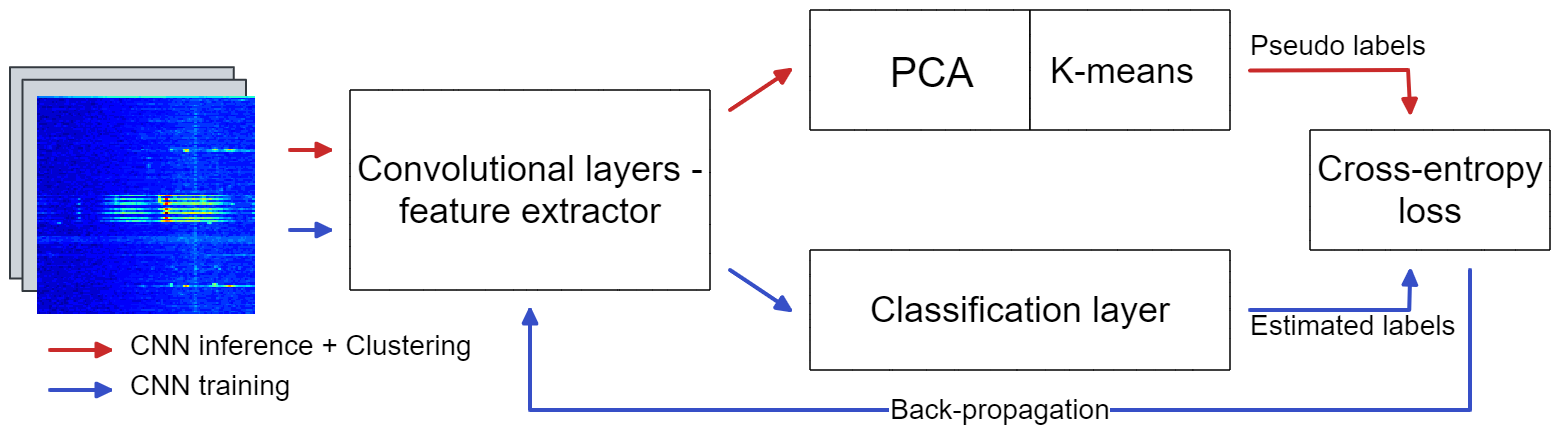}}
    \caption[width=\textwidth]{Feature learning and clustering architecture.}
    \label{fig:FigureArchitectureDiagram}
    \vspace{-6mm}
\end{figure}

The CNN-based feature learning and clustering architecture used in this study is depicted in Figure~\ref{fig:FigureArchitectureDiagram}. The system design contains a representation learning block and a clustering block. The representation learning block contains CNN realized through ResNet18 \cite{he2016deep} followed by a principal component analysis (PCA) method performing automatic learning of reduced dimensionality feature representation. The clustering block then processes the data provided by the representation learning block and provides supervision pseudo-labels in the next training step.

During the iterative training process, the CNN outputs are compared to the pseudo-labels generated by the K-means clustering and the difference is propagated back to guide the training. Clustering and CNN weights training are performed in alternating manner. The procedure stops when the predefined number of iterations (training epochs) is reached.

\section{Methodology} 
\label{sec:methodology}

In this section we propose methodology for interpretation and explanation of the SSDL architecture used for clustering of wireless spectrum activity. Given that the architecture consists of two major building blocks, CNN representation learning and a K-means clustering block, we use separate explainability approaches, each suitable for one of the blocks. The main goal is to methodically analyze the architecture building blocks and provide answers to the questions:
\begin{itemize}
    \item What is the provided output?
    \item Why the architecture provides the obtained outputs?
    \item How the architecture provides the obtained outputs?
\end{itemize}

\subsection{Explaining the CNN - what, why and how}
%
To explain the work of the CNN part of our model, we use the Guided Backpropagation approach proposed in \cite{springenberg2014striving}. The approach calculates back-propagation gradients and alters the activation so that only non-negative gradients are back-propagated. Alteration comes from the idea that positive gradients are features the neuron is interested in, and negative gradients are features that the neuron is not interested in.
With this approach we provide an answer about which part of the spectrograms is contributing to the classification. The results should answer the question if the CNN is "interested" in the right content from the input data, in our case the transmission burst, or the classification is done based on other parts of the spectrograms.

\begin{figure}[htbp]
    \centering
    \subfloat[Cumulative sum of variance ratio.\label{fig:var_ratio}]{%
        \includegraphics[width=0.7\columnwidth]{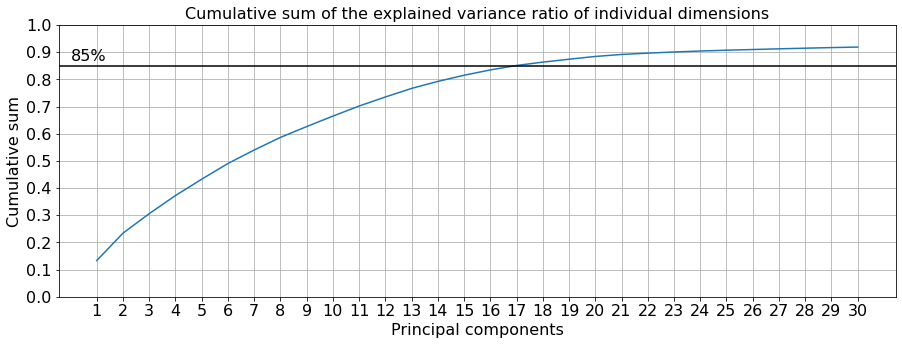}}
    \hfill
    \subfloat[t-SNE on the CNN-produced $1 \times 512$ feature space\label{fig:full_t-SNE}]{%
        \includegraphics[width=0.49\columnwidth]{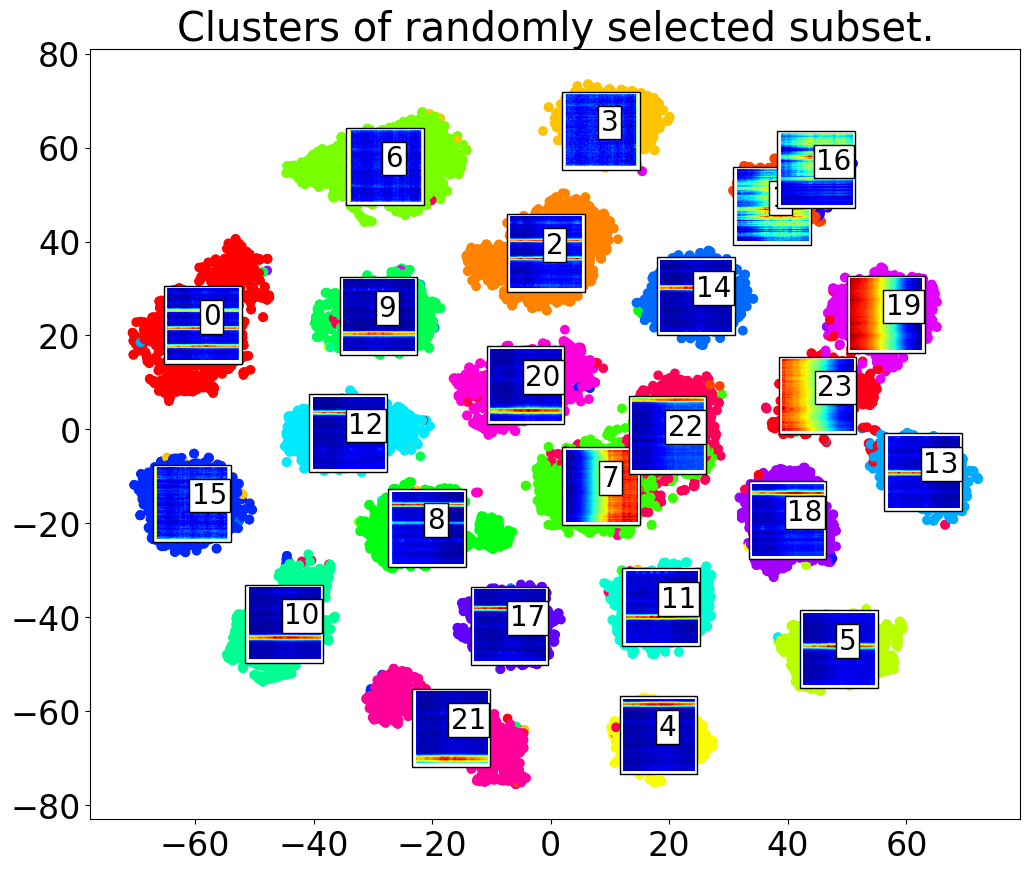}}
    \hfill
    \subfloat[t-SNE on the PCA-transformed, $1 \times 20$ feature space\label{fig:PCA20t-SNE}]{%
        \includegraphics[width=0.49\columnwidth]{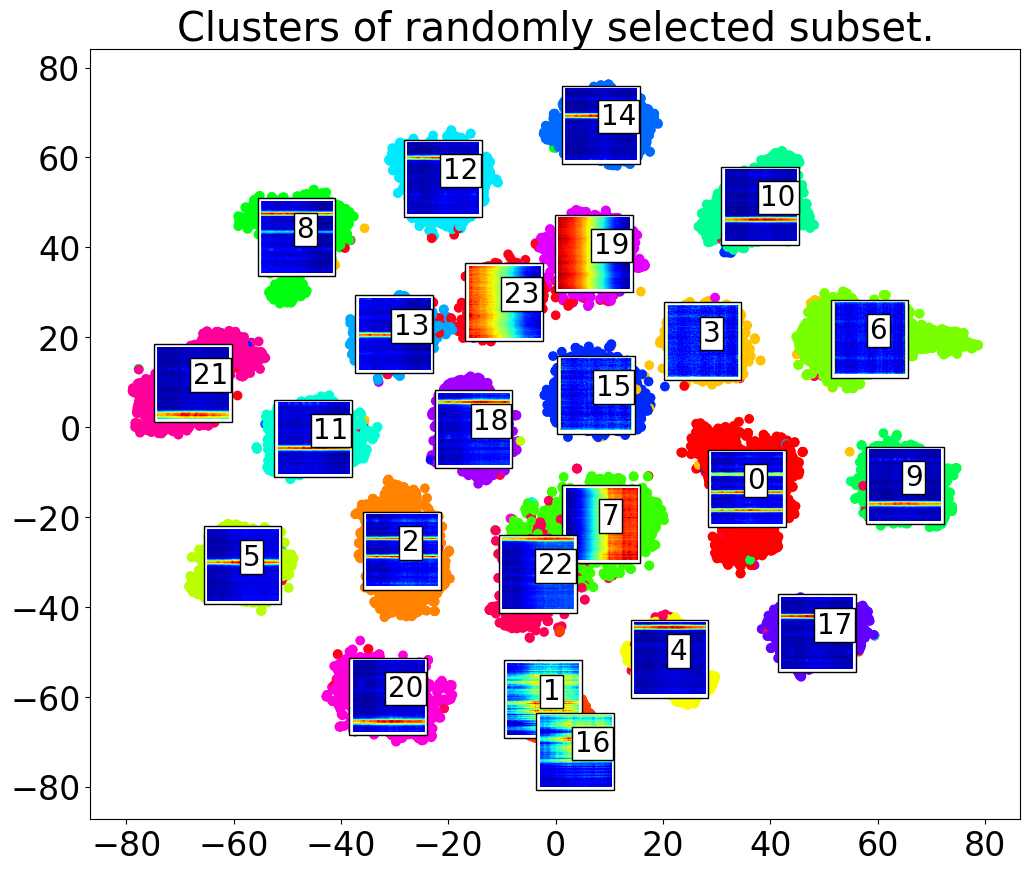}}
\caption{Effect of dimensionality reduction using PCA.}
\label{fig:PCA_dim_reduction}
\vspace{-5mm}
\end{figure}

\subsection{Explaining the K-means - what, why and how}

The clustering algorithm, in our case the K-means, works on the feature space that is provided by the CNN-based feature extractor. Before clustering, the feature space is additionally reduced to a smaller number of dimensions (20 in our experiments) by using the PCA-dimensionality reduction algorithm. The PCA algorithm provides transformed feature space with respect to the original, with the same number of dimensions that are ordered according to the amount of the Explained Variance Ratio (EVR). 
EVR of a single component is the percentage of variance that is attributed to that component from the total variance in the feature space. The goal of working with the transformed space is to keep only the dimensions that contain the highest variation. Utilizing the PCA as part of the SSDL pipeline and selection of the first $N$ dimensions that express the biggest variance ratio is motivated by the following:

\begin{enumerate}[wide, labelwidth=!, labelindent=0pt]
    \item \label{assumption} Wireless transmission activities, appearing as energy bursts in the spectrograms at specific time-frequency locations, cover only a small area (varying, but not more than around 10\%, i.e. 1600 pixels of the spectrogram, relative to its size of $128\times 128=16384$ pixels). Following this, we assume that most of the dimensions, of the original 512, do not contain information that is relevant for the representation of the transmission patterns, but they represent the noise of regions with very weak or without transmission bursts.
    \item Utilizing lower dimensionality space allows for better explainability and reasoning of eventual groups of closely related data samples. This is especially important in our case, since the individual features provided by CNN do not have any explainable meaning, except that they are some values on the coordinate axes.
    \item PCA transformation is linear and preserves the global data structure, identifying parameters that are close in Euclidean distance sense. This makes a common ground with algorithms that work with Euclidean distance metric, such as K-means, making the whole pipeline more transparent.
\end{enumerate}

The number of dimensions $N$ used in the PCA-reduced dimensionality representation is determined by setting a threshold on the amount of EVR that should be expressed by the $N$ number of dimensions. The threshold is determined by exploring the plot of the cumulative sum of the EVR of the PCA-transformed space dimensions. We verify that the PCA-reduced representation contains the relevant information by employing t-SNE \cite{van2008visualizing}. We compare the 2D t-SNE embeddings of the original $1 \times 512$ feature vectors and the PCA-reduced feature vectors. If our assumption that PCA-reduced space of only $20$ dimensions contains the relevant information, originally contained in $512$ dimensions, is true, then the resulting 2D plots of the t-SNE data will contain no significant differences, regarding the existing clusters.

Explaining of the clustering result and the effect of individual dimensions (features) of the PCA-transformed space is performed by three methods:
\begin{enumerate}[wide, labelwidth=!, labelindent=0pt]
    \item Building a decision tree, with the PCA-transformed features and the result of the K-means clustering. The idea is to create a logic that can imitate the clustering, but also provide information about which features are important for particular clusters/types of transmission activities. Special type of an algorithm for building Shallow Tree that optimizes the depth of the tree is used so as to avoid very deep trees that are less explainable, because explained clusters by the built tree depend on bigger number of features.
    \item Visualization of the average of the clustered spectrograms. Such visualizations show shapes of signals that are most frequent in the formed clusters. Averaged spectrogram is chosen as a relevant representation of a single cluster because of the relatively big size of the dataset and the size of the clusters of more than 10,000 samples, which is lowering the effect of outliers on the average images.
    \item Visualization of the histogram of the origin of the samples from each cluster/leaf in the tree. The histogram is calculated by counting the number of samples that fall into each of the band segments, for each of the formed clusters. This visualization explains the connection between the origin frequency of the clustered samples and the features in the PCA-transformed space.
\end{enumerate}

By analysing the results obtained with the three methods, we interpret the relation between the input raw data, the features of the PCA-transformed space and the formed clusters as a result of the K-means clustering. So, for each cluster we provide key features, average spectrogram and the origin frequency of its samples.

\section{Representative results for explainability}
\label{Sec:results}

\subsection{Model development and observations}
As explained in Section~\ref{Sec:architecture}, the model is developed through iterative alternating between the representation learning and the clustering block. The clustering is randomly initialized in each iteration. This means that clusters that are being gradually formed during the iterations are getting different labels in each cycle. Thus, CNN gets only a single pass of the data to adapt to the new labels. For specific cases, this one pass may be enough \cite{caron2018deep}, but spectrograms have far less content compared to the RGB images in the used ImageNet benchmark dataset. Accordingly, more epochs are needed for the CNN to adapt to the formed groups provided by the K-means. This motivated experimentation with the number of epochs after which the clustering is applied in order to provide enough iterations for the CNN to adapt.

\begin{figure}[h!]
\vspace{-5mm}
    \centering
    \subfloat[Clustering cycle of 1 epoch\label{fig:CEloss:1}]{%
        \includegraphics[width=0.49\columnwidth]{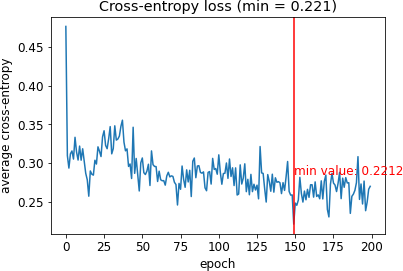}}
    \hfill
    \subfloat[Clustering cycle of 15 epochs\label{fig:CEloss:15}]{%
        \includegraphics[width=0.49\columnwidth]{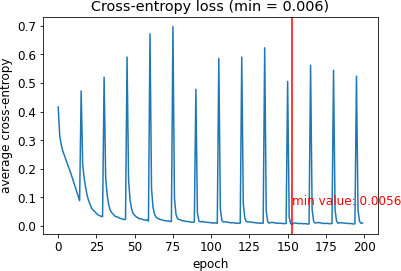}}
    \caption{Cross-entropy loss for different clustering cycles.}
\label{fig:CEloss}
\vspace{-3mm}
\end{figure}

Experiments have shown that 5-15 epochs per clustering cycle provide much better learning curve and significantly lower final loss values. The benefits of using clustering cycle bigger than one are shown in Figure~\ref{fig:CEloss}. Performing clustering in every epoch causes the loss to obtain fast varying values. Although the general shape in Figure~\ref{fig:CEloss:1} has a trend of decreasing, as the epochs progress, there is not much difference between the starting value of $0.3$ to the minimal value of $0.22$ and end-value of $0.27$. On the other side, setting the clustering cycle to $15$ provides loss function with the shape presented in Figure~\ref{fig:CEloss:15}. Each $15$-epoch, when the clustering is performed, exhibits much higher peaks, up to $0.7$, but the value stabilizes very fast, on the right half of the graph in less than $15$ epochs, ending with loss value of around $0.0056$. The convergence is reached in approximately $100$ epochs. This means that CNN is able to adapt to the clusters almost perfectly, and the peaks appear as a result of reassigning the clusters' labels which is resolved very quickly, probably by adapting only in the final, fully connected layer of the network. Recovery of the small loss values after the peaks is happening much faster in the second half of the training epochs, which means that CNN is more capable of extracting the general features needed for the distinguishing between the clusters. This also supports the previous statement that after $100$ epochs, the CNN learned the important features that are key for the clustering and is only adapting to the random labels reassignment of the K-means after each initialization.

The methodology enabling explainability of the SSDL model, described in Section~\ref{sec:methodology}, was applied to an instance of such model developed by using a 20-dimensional PCA-transformed feature space for clustering, 24 clusters for the K-means and the clustering cycle of 15 epochs.


As can be seen in Figure~\ref{fig:var_ratio}, the first 17 dimensions of the PCA-transformed space contain 85\% of the EVR. The PCA components numbered above 20 make very small contribution to the EVR, so we assume they are much less relevant for the clustering. We verify this assumption by exploring the 2D t-SNE embeddings of the 512-dimensional CNN-provided features (Figure~\ref{fig:full_t-SNE}) and the 20-dimensional PCA-transformed features (Figure~\ref{fig:PCA20t-SNE}). For additional clarity, average spectrograms of the samples from each cluster with the appropriate cluster label and color are shown on the plots. We anticipate that the same number of clusters is apparent in both plots. Also, clusters that are close in the $1 \times 512$ space are also close in the $1 \times 20$ (e.g., clusters . which is expected because the PCA-transformation changes the position and orientation of the coordinate system. We conclude that using the 20 biggest EVR, PCA components makes insignificant difference on the clusterability of the samples, while drastically reducing the complexity of the feature space. The reduced-dimensionality space makes it possible to apply tree-building algorithm on top of the clustering results that can be of a tractable size, considering that there are 24 clusters and 20 features.

\vspace{-1mm}
\subsection{CNN model explainability}

The outcome of the Guided Backpropagation is a map of attributions. A positive attributions score (red) means that the input in that particular position positively contributed to the final prediction, and a negative (blue) means the opposite. The magnitude of the attribution score signifies the strength of the contribution. A zero attribution score (white) means no contribution from that particular feature.

\begin{figure}[htbp]
\vspace{-5mm}
    \centering
    \subfloat[Sample \#1\label{fig:gbp:1}]{%
        \includegraphics[width=0.48\linewidth]{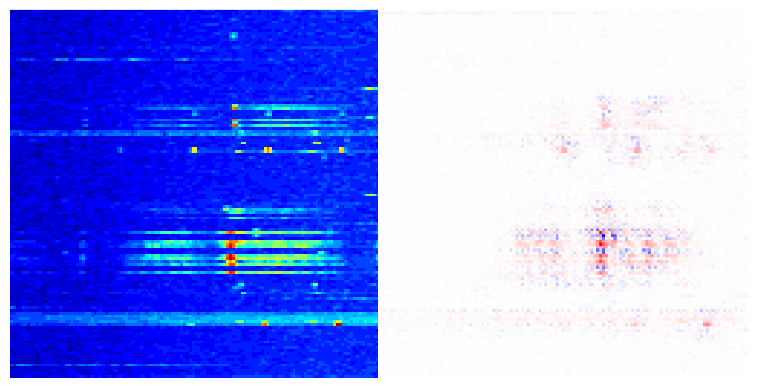}}
    \hfill
    \subfloat[Sample \#2\label{fig:gbp:2}]{%
        \includegraphics[width=0.48\linewidth]{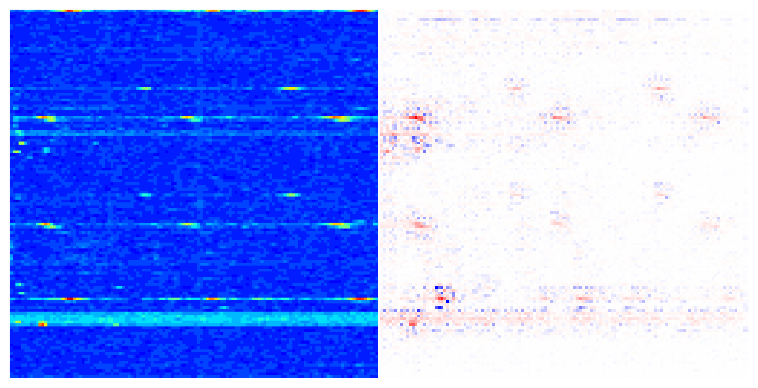}}
    \caption{Spectrum samples (left half) and attribution maps from Guided Backpropagation (right half).}
    \label{fig:gbp}
    \vspace{-4mm}
\end{figure}

In Figure~\ref{fig:gbp}, we present four samples from our spectrum data. Sample~\#1 (Fig.~\ref{fig:gbp:1}) shows that categorization decision is made on two transmissions: a stripe at the bottom, and main transmission near the center of the sample.

\begin{figure}[htbp]
\vspace{-5mm}
    \centering
    \subfloat[Sample \#1\label{fig:avg:1}]{%
        \includegraphics[width=0.48\linewidth]{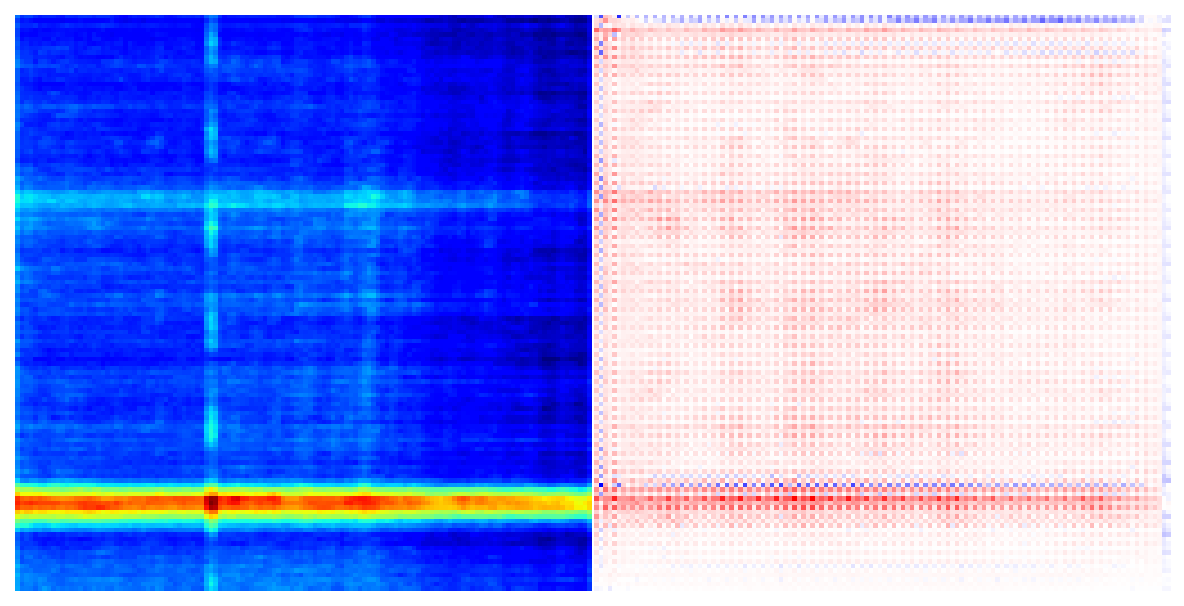}}
    \hfill
    \subfloat[Sample \#2\label{fig:avg:2}]{%
        \includegraphics[width=0.48\linewidth]{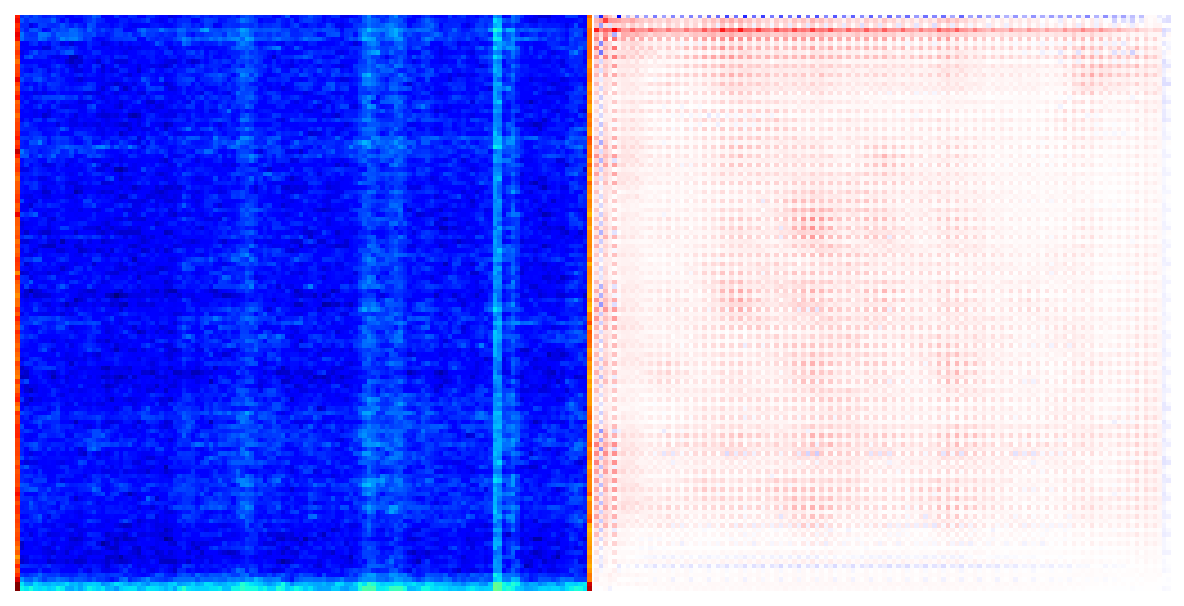}}
    \caption{Average spectrum samples (left half) and average attribution maps from Guided Backpropagation (right half).}
    \label{fig:gbp:avg}

\end{figure}

In Figure~\ref{fig:gbp:avg}, we present average sample and average attribution map of a cluster. Figure~\ref{fig:gbp:avg}a shows that average sample of the selected cluster contains a strong transmission at the bottom, and weak continuous transmissions with different central frequency (vertical stripes).


\begin{figure}[!h]
    \noindent\makebox[\columnwidth]{\includegraphics[width=\columnwidth]{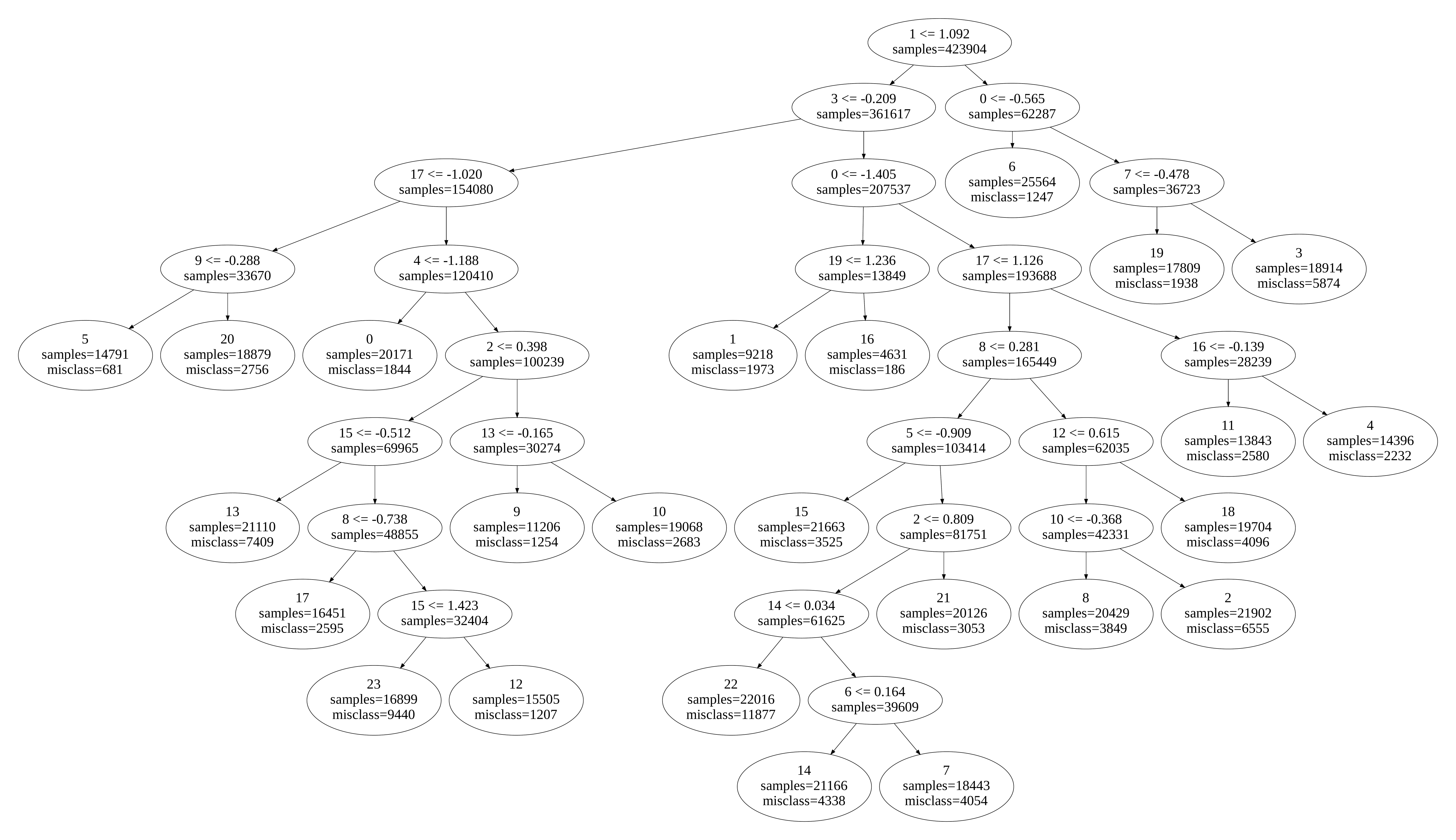}}
    \caption[width=\columnwidth]{Shallow Tree built using the clustering result.}
    \label{fig:ClusTree}
    \vspace{-5mm}
\end{figure}

\begin{figure}[ht]
    \centering
    \noindent\makebox[\columnwidth]{\includegraphics[width=\columnwidth]{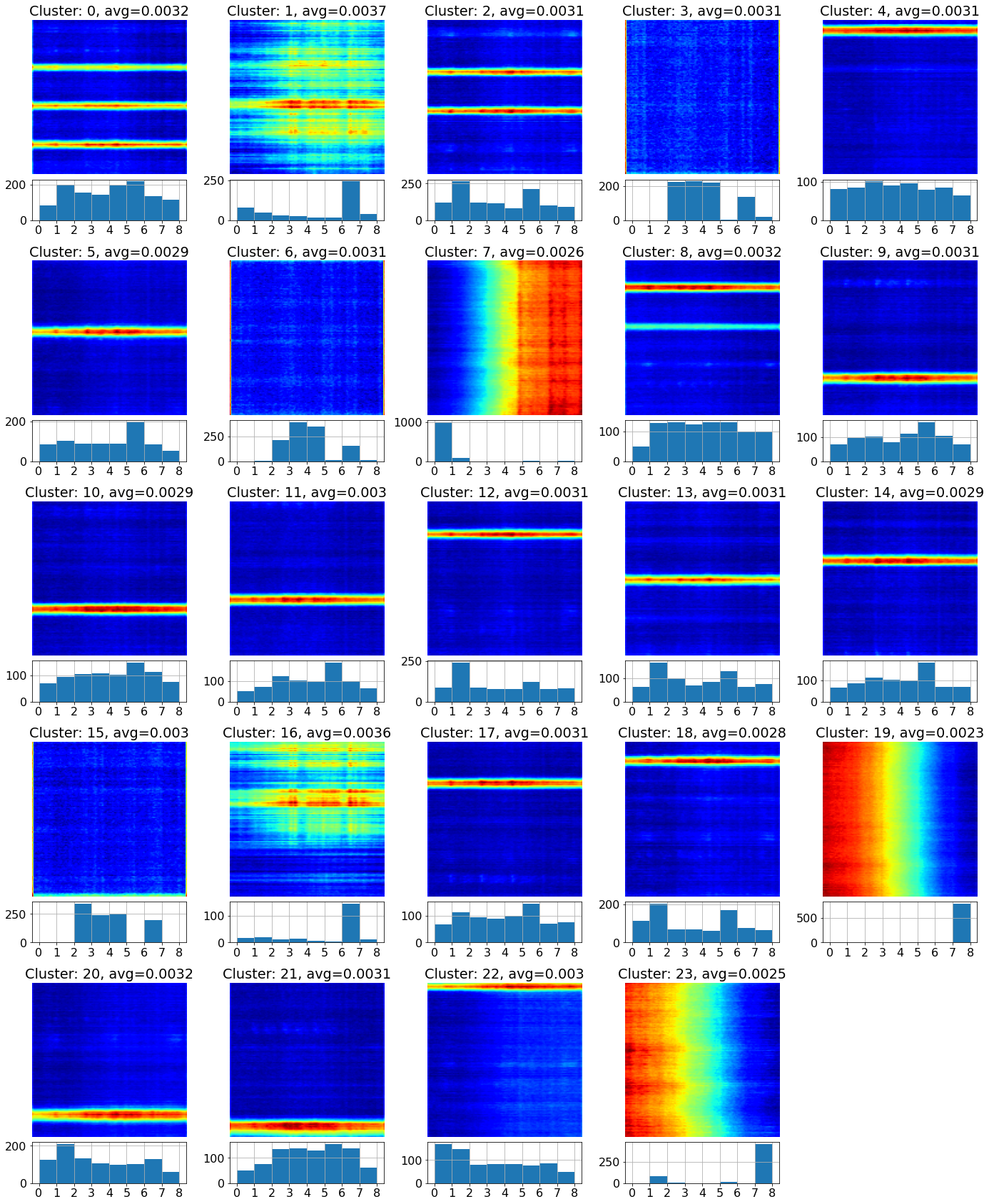}}
    \caption{Averaged spectrograms and histograms of the existing clusters.}
    \label{fig:AvgSpecAndHist}
    \vspace{-6mm}
\end{figure}

\vspace{-1mm}
\subsection{K-means results}
Following the methodology from Section~\ref{sec:methodology}, we built a decision tree based on the features in the PCA-transformed space and the labels provided by the clustering, using an algorithm that optimizes (minimizes) the depth. The only parameter used in the tree building algorithm is the number of leafs/clusters, which means that we are trying to partition the hyperspace in the same number of regions as the number of clusters, 24 in our case. The feature space has $20$ dimensions. The tree developed with these parameters is shown in Figure~\ref{fig:ClusTree}. Nodes of the tree are marked with the feature labels $0-19$ for the $20$ features and the number of samples contained in the branches starting from each node. Leafs of the tree contain the label of the cluster, i.e. $0-23$ for the $24$ existing clusters, and also the total number of samples and the number of falsely classified samples for each of the leafs. The numbers in the nodes and leafs are obtained by inference with the built tree on the complete dataset of $423,904$ samples. 

The tree is complemented with the average spectrogram of each cluster and the histogram of appearance of the samples across the monitored band, shown in Figure~\ref{fig:AvgSpecAndHist}. The averaged spectrograms are calculated using a randomly sampled subset of data. Observation of the average spectrograms in Figure~\ref{fig:AvgSpecAndHist} provides information about the different types of activities that are specific for each cluster and their origin in the band where data is being sensed. 

\vspace{-1mm}
\subsection{Relating features to the specific clusters and content}

By comparing visualizations in Figure~\ref{fig:AvgSpecAndHist} to the decision tree in Figure~\ref{fig:ClusTree}, we discover which features from the PCA-transformed space are important for each of the clustered spectrum activities. The left branch of the tree, distinguished by features 1 and 3, contains spectrum activities (clusters 9, 10, 12, 17, 13, 5, and 20) characterized by horizontal bright stripe along the frequency axis. Histograms of origin of the samples from these clusters have mostly uniform distribution across the entire band. This is expected because the horizontal stripes activities are specific for the IEEE 802.15.4 transmissions. The difference between the clusters is the location of the transmission because of the fixed segmentation of the data into squared, non-overlapping segments, according to Section~\ref{sec:Data}. The only exception in this branch is cluster 23. Quantitatively, this leaf in the decision tree contains the most misclassifications in the branch. Qualitatively, according to the average spectrogram images, it is formed based on the varying background activity, and the corresponding histogram shows that samples come from the right-most part of the band. The high number of misclassifications and the qualitative difference show that the left branch is not suitable for explaining the features that are relevant for the cluster 23.

The small right branch, containing the clusters 6, 19 and 3 shows that features 1, 0 and 7 are important for distinguishing the clusters that do not contain any activity. The difference between 19 and the other two is their frequency origin. While the samples from cluster 19 come from the right-most part of the band, clusters 3 and 6 contain samples mostly from the regions 2-5 according to the corresponding histograms. Two smaller branches are explaining horizontal stripe activities dependent on features 1, 3, 0, 17, 8 (16) which are important for clusters 8, 2 and 18 (4 and 11). Strong activities grouped in clusters 1 and 16 are explained by features 1, 3 and 0. The rest of the clusters are not grouped in significant branches in the tree and each of them is explained by a specific group of features.

Building a decision tree with recalculation of the PCA-representation for explaining the clustering results in the exact same tree which means that the relation between the content of the input spectrograms and the features is consistent. There are only a few tens of samples that are being reassigned across the tree, which is negligible compared to the clusters size of 5-40 thousands. Rerunning the K-means clustering on the PCA-transformed features also makes no changes in building the tree. The only thing that is being changed are the labels of clusters resulting from the random initialization of the K-means clustering. This means that the CNN-provided features, which are then PCA-transformed, form a well distinguished clusters in the 20-dimensional hyperspace.


\section{Conclusions}
\label{Sec:conclusions}

In this paper we proposed a technique for explaining SSDL clustering architectures based on Guided Backpropagation, Shallow Tree building and field-specific visualizations relevant for the wireless communications domain. Using the proposed technique, a clear connection can be determined and interpreted between the important content of the input data, the encoded features and the output classes provided by the clustering algorithm. We demonstrate the effectiveness of our approach by using it on a SSDL model utilized for clustering of wireless transmissions, using unlabelled real-world data.

\section*{Acknowledgments}
This work was funded in part by the Slovenian Research Agency under the grant P2-0016 and in part from the European Union’s Horizon Europe Framework Programme under grant agreement No. 101096456 (NANCY). The NANCY project is supported by the Smart Networks and Services Joint Undertaking and its members.

\bibliographystyle{ieeetr}
\bibliography{biblio}

\end{document}